# Large Language Model-Aided Evolutionary Search for Constrained Multiobjective Optimization

Zeyi Wang[1], Songbai Liu[1], Jianyong Chen[1], and Kay Chen Tan[2]

[1] College of Computer and Software Engineering, Shenzhen University, Shenzhen, China
[2] Department of Computing, The Hong Kong Polytechnic University, Hong Kong SAR
wangzeyi2022@email.szu.edu.com, songbai@szu.edu.cn

**Abstract.** Evolutionary algorithms excel in solving complex optimization problems, especially those with multiple objectives. However, their stochastic nature can sometimes hinder rapid convergence to the global optima, particularly in scenarios involving constraints. In this study, we employ a large language model (LLM) to enhance evolutionary search for solving constrained multiobjective optimization problems. Our aim is to speed up the convergence of the evolutionary population. To achieve this, we finetune the LLM through tailored prompt engineering, integrating information concerning both objective values and constraint violations of solutions. This process enables the LLM to grasp the relationship between well-performing and poorly performing solutions based on the provided input data. Solution's quality is assessed based on their constraint violations and objective-based performance. By leveraging the refined LLM, it can be used as a search operator to generate superior-quality solutions. Experimental evaluations across various test benchmarks illustrate that LLM-aided evolutionary search can significantly accelerate the population's convergence speed and stands out competitively against cutting-edge evolutionary algorithms.

## 1 Introduction

Constrained multiobjective optimization problems (CMOPs) stand out as a specific subset of multiobjective optimization problems (MOPs), distinguished by constraints imposed on objective or variable spaces. These constraints are prevalent in real-world scenarios, ranging from resource limitations [1] to safety requirements [2]. The fundamental challenge in solving CMOPs lies in optimizing multiple conflicting objectives while adhering to these constraints, aiming to identify feasible Pareto optimal solutions. Evolutionary algorithms (EAs), inspired by the mechanisms of species evolution and developed through the simulation of biological evolutionary processes, have emerged as powerful tools for solving MOPs across various domains, including path planning [3, 4] and robotic grasping [5]. Thus, the advancement of specialized constrained multiobjective EAs (CMOEAs) tailored for CMOPs has garnered significant attention. CMOEAs typically strive towards optimal solutions by designing appropriate search operators and constraint handling techniques (CHTs). Despite their potential, existing CMOEAs often suffer from a notable drawback: insufficient convergence speed. This limitation necessitates substantial computational resources (e.g., function evaluations) to effectively solve CMOPs. Addressing this challenge is crucial for enhancing the practical applicability of CMOEAs in solving real-world optimization problems.

To enhance the convergence rate, there has been a growing interest in leveraging machine learning methods to bolster the efficiency of evolutionary search operators [6, 7]. One notable way involves training neural network models to glean heuristic insights from existing solutions, effectively serving as learnable search operators to generate higher-quality solutions [8]. This learnable evolutionary search strategy has shown promise in expediting the optimization of MOPs with many objectives and large-scale decision spaces [9]. However, its application to solving CMOPs remains largely unexplored. Furthermore, existing studies have primarily utilized relatively simplistic models, often capable of only acquiring shallow knowledge. Consequently, these models fail to deliver substantial acceleration effects on evolutionary search processes. Thus, there is a pressing need for more sophisticated models capable of capturing deeper insights to augment the effectiveness of evolutionary search operators.

Recently, large language models (LLMs) have undergone rapid advancements, leading to significant breakthroughs across various research domains. Leveraging vast amounts of data, LLMs have demonstrated impressive capabilities in reasoning and prediction. This adaptability enables LLMs to generate targeted solutions for a multitude of optimization problems through tailored prompt engineering, showcasing exceptional generalization performance. In this context, researchers have explored the potential of integrating LLMs with evolutionary computation techniques. This fusion of LLMs with EAs presents an promising avenue for advancing optimization methodologies [10, 11]. By incorporating the strengths of both LLMs and EAs, researchers aim to enhance the efficiency and effectiveness of optimization processes. This synergistic approach capitalizes on the deep understanding and adaptability offered by LLMs, potentially leading to significant advancements in solving complex optimization challenges, e.g., CMOPs.

Inspired by these developments, our research investigates the fusion of LLMs with CMOEAs to tackle CMOPs. We propose a novel approach wherein a LLM is integrated with traditional evolutionary search operators to collaboratively generate offspring solutions. At each generation, solutions from the evolutionary population are passed to the LLM. By employing tailored prompt learning techniques, we guide the LLM to refine its outputs, aiming to produce offspring solutions with improved convergence characteristics. Our main contributions can be summarized as follows:

- We devise a specialized prompt learning mechanism to fine-tune the LLM, incorporating considerations of both constraint violations and objective values in solving CMOPs. In this way, the LLM can generate solutions that excel in both objective-based performance and constraint adherence.
- We explore the potential of integrating the refined LLM as a search operator within a classical CMOEA framework. This integration involves working alongside traditional evolutionary search operators, all aimed at collectively accelerating the convergence rate of the population.

The structure of this paper is as follows: Section 2 introduces CMOPs and LLMs, discussing the rationale behind our study. Section 3 details the integration of LLMs within CMOEAs. Section 4 presents experimental validation. Finally, Section 5 summarizes the findings and suggests our future research directions.

## 2    Related work and Motivation

### 2.1    Constrained Multiobjective Optimization Problem

In general, a CMOP can be defined as follows:

$$\text{Minimize } F(x) = (f_1(x), \ldots, f_m(x))$$
$$s.t. \begin{cases} g_i(x) \leq 0, & i = 1, \ldots, q \\ h_i(x) = 0, & i = q+1, \ldots, l \end{cases} \quad (1)$$

where $x$ is a decision variable with $n$ dimensions, $F(x)$ represents the set of $m$ objective functions to be optimized. The functions $g(x)$ and $h(x)$ represent the $q$ inequality constraints and $l - q$ equality constraints, respectively. Furthermore, the assessment of feasibility for a solution $x$ is typically based on its constraint violation degree $CV(x)$, which is defined as follows:

$$CV(x) = \sum_{i=1}^{q} \max(0, g_i(x)) + \sum_{i=q+1}^{l} \max(0, |h_i(x) - \delta|) \quad (2)$$

where $\delta$ is a boundary relaxation parameter for the purpose of transforming equality constraints into inequality constraints. Typically, $\delta$ is a very small positive value.

CMOEAs solve the diverse constraint challenges within CMOPs by developing specific CHTs. Common CHTs encompass five main strategies: penalty-based [12], dominance-based [13], transformation-based [14], multi-stage [15], and multi-population [16] methods. The penalty-based method incorporates $CV(x)$ as a penalty factor into the objective function. The dominance-based method sorts solutions according to specific dominance relations, such as constrained dominance principle (CDP). The transformation-based method converts CMOPs into MOPs, treating constraints as additional optimization objectives. The multi-stage method divides the optimization process into several stages, each dedicated to achieving specific optimization goals. The multi-population method establishes multiple populations or archives to tackle different optimization tasks derived from the original task.

However, the effectiveness of CHTs is contingent upon the characteristics of the problem at hand and does not inherently impact the convergence speed of the population. Conversely, the convergence speed of CMOEAs is primarily determined by the choice and efficacy of the search operators, such as crossover [17], mutation, and differential evolution (DE) [18] operators. Researchers have consistently dedicated efforts to enhance algorithmic performance either by developing more efficient search operators or by employing a combination of multiple search operators. IMTCMO utilizes two variants of DE operators, DE/rand/1 and DE/current-to-best/1, to bolster the algorithm's diversity and convergence, respectively. Similarly, DBEMTO [19] incorporates both genetic algorithm (GA) operators and DE operators to achieve a more balanced search in terms of diversity and convergence. Furthermore, PKAEO [20] aims to adaptively select appropriate search strategies by constructing a mapping model between population states and search strategies. LCMOEA [21] attempts to explore new solution spaces by generating offspring through the training of a deep learning network.

### 2.2   Large Language Models

LLMs represent a class of large-scale deep neural networks built upon the transformer architecture, often comprising billions to tens of trillions of parameters. These models are trained using vast datasets to predict given inputs effectively. Initially, LLMs segment input data into smaller units known as tokens, leveraging deep neural networks to encode and decode complex relationships and patterns among these tokens. This process enables accurate prediction and inference of results. Over the past few years, the rapid expansion of model scale and the availability of extensive training datasets have accelerated the development of LLMs, resulting in significant improvements in predictive accuracy. This growth has led to an expanding research landscape centered around LLMs, paving the way for new interdisciplinary research directions.

Recent research trends demonstrate a growing interest in exploring the potential applications of LLMs in optimization [22, 23]. These studies generally fall into two categories: one involves using LLMs directly as black-box tools for addressing optimization problems [24], while the other focuses on integrating LLMs with evolutionary computation to explore their potential as genetic operators within the EA framework. For instance, the recently proposed LMEA aims to utilize LLMs for generating offspring to solve the traveling salesman problem (TSP) [11]. Additionally, researchers have integrated LLMs into the classical MOEA/D framework [25] to generate offspring and have explored the feasibility of creating new operators based on offspring generated by LLMs [10]. Comparative analyses have shown that EAs and LLMs demonstrate similar performance in multiple key aspects, indicating that LLMs can effectively simulate evolutionary processes [26]. These findings show the potential of using LLMs for automated algorithm design in optimization. While some progress has been made [27, 28], the integration of LLMs in optimization is still in its early stages, leaving plenty of room for further exploration and development.

### 2.3   Motivation

The challenge faced by CMOEAs lies in effectively identifying a feasible solution set that closely approximates the constrained pareto front (CPF). This task is particularly difficult due to the presence of constraints, which can result in decision or objective spaces being covered by infeasible regions, often extensive and continuous. Navigating through these regions poses a significant challenge to achieving population convergence, often requiring substantial computational resources (e.g., hundreds of thousands of function evaluations). While previous research has proposed various search strategies to address this issue, the emergence of LLMs offers new possibilities. Through the formulation of promising prompts, LLMs can potentially be harnessed to generate offspring, presenting a promising avenue for enhancing population convergence. Despite prior attempts to integrate LLMs into EAs, significant success in this regard has yet to be achieved. This study aims to address this gap by exploring the use of LLMs for offspring generation with the goal of accelerating population convergence in solving CMOPs. The subsequent sections will delve into the strategies employed for utilizing LLMs and will present experimental results to validate the effectiveness.

## 3 The Proposed Algorithm

This section provides a detailed explanation of the proposed solver, termed CMOEA-LLM, which combines an existing CMOEA with a LLM to enhance optimization in solving CMOPs. We start by introducing the general framework of CMOEA-LLM. Following that, we explore how the LLM collaborates with evolutionary search operators. Subsequently, to address the unique challenges of CMOPs, we introduce a specialized prompt aimed at guiding the LLM to generate novel and efficient solutions that achieve superior objective-based performance and constraint satisfaction.

---
**Algorithm 1**: General Framework of the proposed CMOEA-LLM

**Input:** the target CMOP, population size $N$, maximum function evaluations $FE_{max}$
**Output:** the final population $P$
1: initialize the population $P$ with $N$ random solutions;
2: initialize the function evaluation counter $FE = 0$;
3: **while** $FE <= FE_{max}$ **do**
4:     $O^e \leftarrow$ generate $0.9N$ solutions by the evolutionary search in the used CMOEA;
5:     $O^l \leftarrow$ generate $0.1N$ solutions by the proposed LLM-Aided search;
6:     $U \leftarrow P \cup O^e \cup O^l$, then set $P$, $O^e$, and $O^l$ as empty populations;
7:     $P \leftarrow$ select $N$ solutions from $U$ by the environmental selection in the CMOEA;
8:     $FE = FE + N$;
9: **end while**
10: **return** the final population $P$

---

### 3.1 The General Framework of CMOEA-LLM

**Algorithm 1** outlines the general framework of CMOEA-LLM, which requires inputs including the target CMOP to be solved, the population size ($N$), and the maximum function evaluation budget ($FE_{max}$). First, the evolutionary population $P$ is initialized with $N$ random solutions, and the function evaluation counter $FE$ is set to 0. The algorithm then enters a main loop that continues until a predefined termination condition is met, i.e., $FE > FE_{max}$. During each iteration of the loop, two offspring populations are generated: $O^e$, consisting of 90% of the solutions, using the specific evolutionary search method adopted by the CMOEA; and $O^l$, comprising 10% of the solutions, generated by the LLM, which is finetuned by a specifically designed prompt engineering process (details are provided in Section 3.3). Afterward, all solutions from populations $P$, $O^e$, and $O^l$ are combined into the union population $U$. Finally, the population $P$ is updated using environmental selection, as proposed in the adopted CMOEA. The algorithm iterates until the termination condition is met, at which point the final population $P$ is returned as the output result.

It's important to note that the choice of CMOEA can be adapted to suit the specific requirements of the CMOP, aiming to achieve superior optimization performance. In our study, we opt for CCMO [29] as the foundational CMOEA for investigation. CCMO is a classical multi-population coevolutionary framework, evolving one population for the CMOP while considering all constraints and another population while ignoring all constraints. This framework offers flexibility in adjusting evolutionary search and environmental selection strategies to suit the specific CMOP at hand. Therefore, our study builds upon the CCMO framework, modifying its search strategies by integrating LLMs with traditional evolutionary search operators to collaboratively generate offspring solutions. This proposed solver, named CCMO-LLM, is designed to enhance both the efficiency and quality of solution generation.

### 3.2   Reproduction with LLM-Aided Search

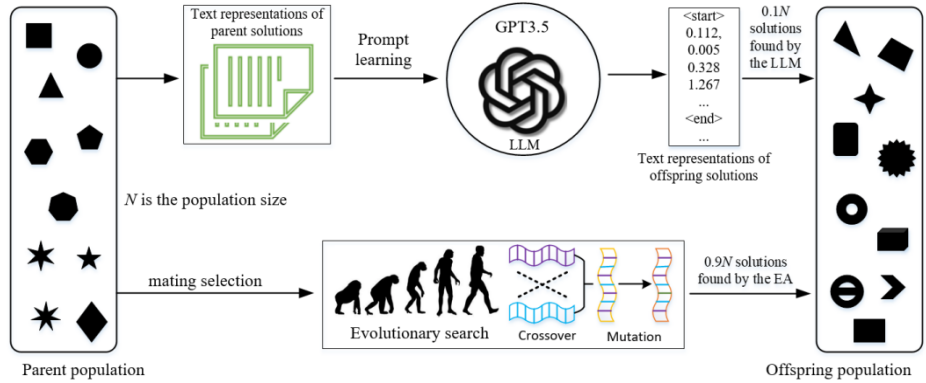

**Fig. 1.** Illustrating the integration of a LLM and an evolutionary search for reproduction.

To overcome the challenges posed by token length limitations and runtime constraints of LLMs, we adopt a hybrid strategy. This strategy involves harnessing a LLM to generate a limited number (i.e., 10%) of solutions with promising convergence in each generation, while the majority (i.e., 90%) of solutions are still produced using traditional evolutionary search operators. Specifically, we integrate GA operators with LLMs for offspring generation within the CCMO framework. This entails each of the two populations in CCMO-LLM generating 5% of offspring solutions through the LLM. To ensure the LLM receives adequate input solutions and that this process aligns with the input-output ratio of GA operators, we randomly select 10% of solutions from each population as inputs for the LLM. This method aims to maximize the capabilities of the LLM while navigating their limitations, ensuring algorithmic efficiency and effectiveness. This hybrid approach aims to maximize the capabilities of the LLM while mitigating its limitations, thus ensuring both algorithmic efficiency and effectiveness. For a detailed illustration of how LLMs are integrated with evolutionary search operators to generate offspring solutions, refer to Fig. 1.

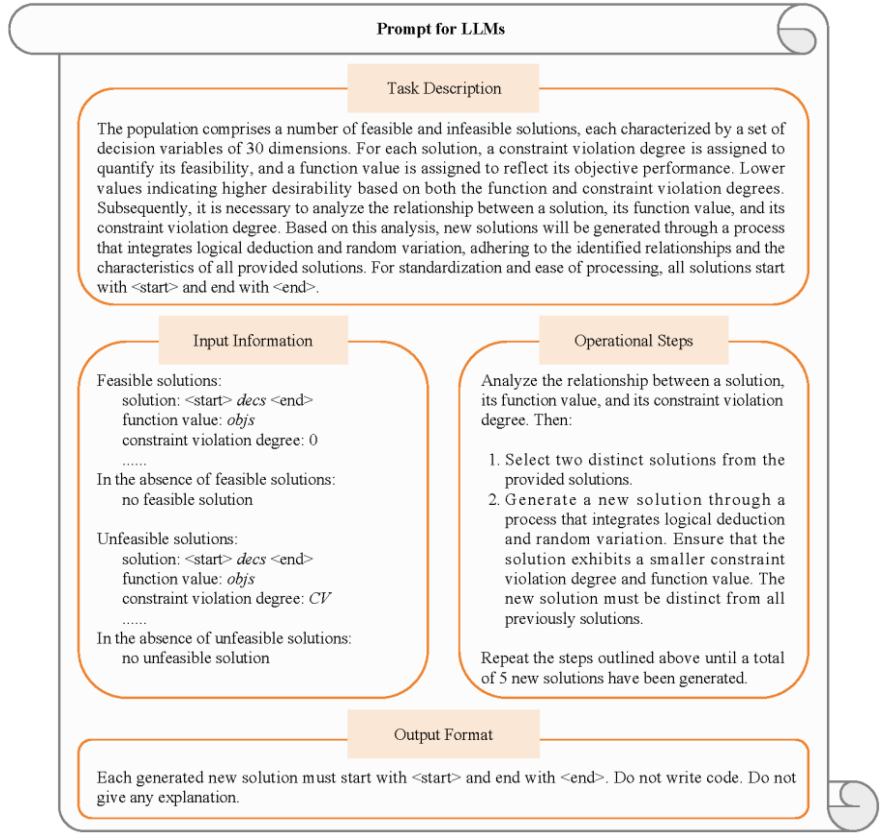

**Fig. 2.** An example of a prompt to guide the LLM for generating promising solutions.

### 3.3   Prompt Engineering for the LLM

In CCMO-LLM, the LLM serves as an innovative search strategy for generating new solutions. It's crucial to recognize that the LLM relies on prompts to perform specific tasks, making the generation of solutions essentially a form of prompt engineering. In addressing CMOPs, we craft a prompt tailored for generating offspring using the LLM, with a focus on improving both objective-based performance and constraint violation degree. These prompts typically consist of four main components:
- Task Description: this part outlines the tasks to be accomplished and provides a brief overview of the input information.

- Input Information: this part offers comprehensive details regarding the solution input, ensuring all relevant information is provided.
- Operational Steps: here, a detailed explanation of the specific steps employed by LLM during the offspring generation process is provided, ensuring clarity, and understanding.
- Output Format: this section stipulates the output format for the LLM to facilitate the extraction and evaluation of the generated results.

Fig. 2 illustrates an example prompt structured as follows: 1) **Task Description**: This part begins by clarifying the information regarding the solution to be input into the LLM, along with establishing the hierarchical relationship between objective-based performances (indicated by function value) and constraint violation degree. Subsequently, the task expected to be accomplished by the LLM is defined. 2) **Input Information**: detailed contents of both feasible and infeasible solutions are sequentially provided, including decision variables (*decs*), objective values (*objs*), and constraint violation degrees (*CV*) of the solutions. 3) **Operational Steps**: the LLM is tasked with selecting two solutions from the provided inputs and generating a completely new solution based on them. However, explicit guidance on how to select and generate solutions is not provided. 4) **Output Format**: the output format for the LLM is strictly defined, requiring the output to be placed between <start> and <end> tags and excluding any additional explanations. This design facilitates the identification and evaluation of the returned solutions. This prompt design significantly reduces reliance on expert knowledge in the field of CMOPs, encompassing only a fraction of professional knowledge.

## 4   Experiments and Results

### 4.1   Experiment Settings

To evaluate the performance of CCMO-LLM comprehensively, we employ the DASCMOP test suite, which presents challenges in convergence, diversity, and feasibility through three distinct types of constraint functions. The first type of constraint function divides the Pareto Front (PF) of a CMOP, thereby posing a diversity challenge; the second type regulates the ratio of feasible regions within the search space, presenting a feasibility challenge; and the third type constrains the attainable boundaries of objectives, introducing a convergence challenge.

Furthermore, we select the original CCMO algorithm and five other classical algorithms as comparative algorithms to assess the effectiveness of LLM-generated solutions. These five comparative algorithms can be categorized into two groups based on their population structure: single-population-based algorithms (including c-DPEA [30], CMME [31]) and multi-population-based algorithms (including MCCMO [32], MTCMO [33], and CCMO). To ensure fair comparisons, we set the parameters of the comparative algorithms according to their original papers.

For a comprehensive evaluation, we utilize two established metrics: inverted generational distance (IGD [34]) and hypervolume (HV [35]). These metrics collectively assess the convergence and diversity of the resulting populations. The population size (*N*) for all algorithms is fixed at 100, while the maximum function evaluations ($FE_{max}$) are set to 10,000. In the CCMO-LLM, we utilize the GPT-3.5 Turbo model as the LLM optimizer. To ensure statistical significance, each algorithm is executed 10 times independently, and the Wilcoxon test, with a significance level of 0.05, is employed to ascertain the statistical significance of differences among the algorithms. In the results, "+", "-", or "=" respectively indicate whether the compared algorithm exhibits better, worse, or similar performance compared to CCMO-LLM. All experiments were conducted on the PlatEMO platform, which is also used in [30]-[33].

### 4.2    Results on the DASCMOP test suite

**Table 1.** The average IGD results obtained by CCMO-LLM and its five competitors on the DASCMOP test suite (termed CMOP1 to CMOP9 problems).

| Problem | c-DPEA | CMME | MCCMO | MTCMO | CCMO | CCMO-LLM |
|---|---|---|---|---|---|---|
| CMOP1 | 7.3863e-1 (3.72e-2) = | 7.2262e-1 (3.14e-2) = | **1.5884e-1 (2.38e-1) +** | 7.6587e-1 (3.51e-2) - | 7.2208e-1 (4.06e-2) = | 6.9009e-1 (5.22e-2) |
| CMOP2 | 2.5708e-1 (1.86e-2) - | 2.6934e-1 (1.95e-2) - | **1.0081e-1 (1.10e-1) +** | 2.7906e-1 (1.94e-2) - | 2.8762e-1 (2.48e-2) - | 2.2896e-1 (3.31e-2) |
| CMOP3 | 3.4465e-1 (2.93e-3) = | 3.5551e-1 (3.20e-2) = | **1.6945e-1 (1.43e-1) +** | 3.0016e-1 (4.86e-2) + | 3.4728e-1 (4.43e-2) = | 3.4555e-1 (1.91e-3) |
| CMOP4 | 6.5232e-1 (1.81e-2) = | 7.8883e-1 (3.12e-1) = | 8.8332e-1 (1.61e-1) = | 1.0073e+0 (1.72e-1) - | 1.1100e+0 (1.12e-1) - | **4.1383e-1 (1.91e-1)** |
| CMOP5 | 8.0831e-1 (1.63e-1) - | 9.0770e-1 (4.89e-2) - | 5.6340e-1 (0.00e+0) = | 8.8604e-1 (7.01e-2) - | 8.2912e-1 (4.37e-2) - | **1.8723e-1 (2.24e-1)** |
| CMOP6 | 8.9820e-1 (2.43e-1) - | 8.8363e-1 (1.69e-2) - | 8.9001e-1 (4.19e-1) - | 9.6849e-1 (3.13e-1) - | 8.6799e-1 (1.60e-1) - | **3.1683e-1 (2.44e-1)** |
| CMOP7 | 1.0770e+0 (2.16e-1) - | 9.3626e-1 (2.15e-1) - | 6.6536e-1 (4.05e-1) - | 1.1453e+0 (2.19e-1) - | 9.6444e-1 (2.24e-1) - | **2.8787e-1 (2.69e-1)** |
| CMOP8 | 1.1753e+0 (1.19e-1) = | 1.0865e+0 (2.48e-1) - | 7.9294e-1 (1.27e-1) = | 1.1239e+0 (1.48e-1) - | 1.1478e+0 (1.93e-1) - | **4.7771e-1 (3.52e-1)** |
| CMOP9 | 3.6567e-1 (8.19e-2) - | 4.6980e-1 (8.45e-2) - | **1.7877e-1 (1.50e-1) =** | 4.6272e-1 (6.22e-2) - | 4.7666e-1 (6.86e-2) - | 2.9173e-1 (3.90e-2) |
| +/-/= | 0/5/4 | 0/6/3 | 3/2/4 | 1/8/0 | 0/7/2 | ---- |

Tables 1 and 2 respectively present a comparative performance analysis of CCMO-LLM and its five competitors based on average IGD and HV values. In these tables, the best results for each tested CMOP are highlighted in bold, and if an algorithm fails to find any feasible solution, the corresponding result is marked as "NaN".

The analysis of these results clearly demonstrates that the introduction of LLM for offspring generation significantly enhances the performance of CCMO. Compared to the original CCMO algorithm, there is a noticeable improvement in most of the nine problems, especially DASCMOP4-8. This improvement validates the good conver-gence properties of the solutions generated by LLM, facilitating faster initial convergence of the population. Moreover, when compared with four other algorithms, CCMO-LLM exhibits strong competitiveness. Except for MCCMO, CCMO-LLM outperforms the other algorithms in five or more of the nine problems. Particularly noteworthy is the performance on DASCMOP4 and DASCMOP8, where three of the five comparative algorithms fail to find feasible solutions within the evaluation limit, whereas CCMO-LLM consistently finds feasible solutions and achieves the best results. This further demonstrates that offspring generated by LLMs can accelerate population convergence. Furthermore, CCMO-LLM demonstrates superior performance in the DASCMOP7-9, suggesting LLM's potential in tackling many-objective optimization problems.

**Table 2.** The average HV results obtained by CCMO-LLM and its five competitors on the DASCMOP test suite (termed CMOP1 to CMOP9 problems).

| Problem | CMME | EMCMO | MCCMO | MTCMO | CCMO | CCMO-LLM |
|---|---|---|---|---|---|---|
| CMOP1 | 3.7676e-2 (2.15e-2) - | 6.1168e-2 (4.69e-2) = | **2.8115e-1 (1.12e-1)** + | 1.8286e-2 (2.38e-2) - | 4.6824e-2 (9.95e-3) = | 5.4213e-2 (2.16e-2) |
| CMOP2 | 3.6041e-1 (7.43e-3) = | 3.5980e-1 (9.57e-3) + | **4.4552e-1 (6.18e-2)** + | 3.5340e-1 (6.17e-3) = | 3.5489e-1 (1.10e-2) = | 3.5229e-1 (1.41e-2) |
| CMOP3 | 3.0916e-1 (6.01e-4) = | 3.0854e-1 (1.70e-3) = | **3.7784e-1 (6.31e-2)** = | 3.2584e-1 (2.24e-2) = | 3.1431e-1 (1.64e-2) = | 3.0926e-1 (4.39e-4) |
| CMOP4 | 0.0000e+0 (0.00e+0) = | 9.5649e-3 (1.66e-2) - | 2.0039e-2 (2.83e-2) = | 0.0000e+0 (0.00e+0) - | 0.0000e+0 (0.00e+0) - | **1.5856e-1 (8.13e-2)** |
| CMOP5 | 4.0386e-2 (3.28e-2) - | 2.1813e-2 (1.93e-2) - | 1.2006e-1 (0.00e+0) = | 3.5238e-2 (3.16e-2) - | 3.6376e-2 (3.15e-2) - | **3.6588e-1 (1.32e-1)** |
| CMOP6 | 2.3124e-2 (2.12e-2) - | 3.4331e-2 (8.19e-3) - | 6.9803e-2 (1.40e-1) - | 3.3638e-2 (3.14e-2) - | 1.2380e-2 (2.14e-2) - | **2.7685e-1 (1.33e-1)** |
| CMOP7 | 2.0542e-2 (2.44e-2) - | 4.2227e-2 (3.70e-2) - | 1.3109e-1 (1.07e-1) - | 1.4334e-2 (1.75e-2) - | 4.0092e-2 (3.43e-2) - | **2.8576e-1 (1.01e-1)** |
| CMOP8 | 0.0000e+0 (0.00e+0) - | 0.0000e+0 (0.00e+0) - | 6.5199e-2 (4.47e-2) = | 0.0000e+0 (0.00e+0) - | 2.0105e-3 (4.50e-3) - | **1.4272e-1 (8.90e-2)** |
| CMOP9 | 2.3958e-1 (2.03e-2) = | 2.1142e-1 (2.39e-2) - | **3.1829e-1 (5.51e-2)** + | 2.0016e-1 (2.25e-2) - | 2.0684e-1 (1.90e-2) - | 2.3475e-1 (2.00e-2) |
| +/-/= | 0/5/4 | 1/6/2 | 3/2/4 | 0/7/2 | 0/6/3 | ----- |

To reflect the comparative results more comprehensively, we employ the Friedman test to rank the performance of CCMO-LLM and its competitors based on the IGD and HV metrics, as shown in Fig. 3. A lower average ranking indicates superior algorithm performance. The findings demonstrate that, although CCMO-LLM ranks slightly below MCCMO on the HV metric, it achieves the highest ranking on the IGD metric. This highlights the effectiveness of combining LLMs with GA operators to improve the convergence rate of the population.

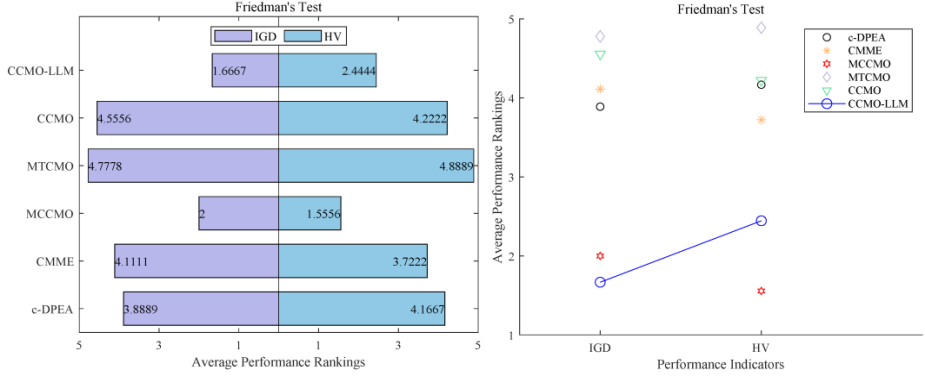

**Fig. 3.** The rankings of Friedman's test obtained by CCMO-LLM and comparison algorithms on DASCMOP test suites respectively over 10 runs.

Further, to validate whether the offspring generated by LLM indeed accelerate the convergence of the population, we display the final solutions obtained by CCMO and CCMO-LLM in solving the DASCMOP test suite in Figs. 4-6. The results in Fig. 5 indicate that for DASCMOP4-6, CCMO-LLM, with LLM integration, exhibits higher convergence, successfully navigating through infeasible regions. For DASCMOP7-9 in Fig. 6, CCMO-LLM achieves a better diversity distribution, suggesting that the introduction of LLM not only aids in accelerating convergence but also contributes to maintaining solution diversity.

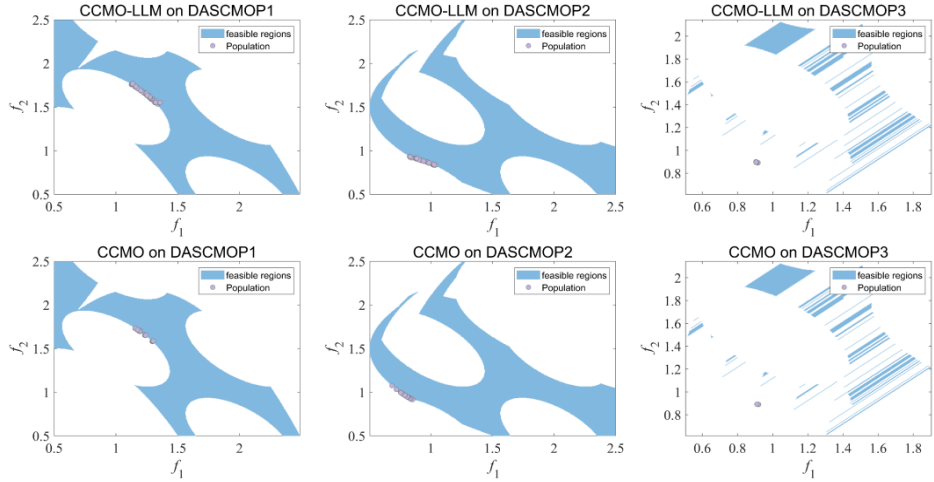

**Fig. 4.** The final population of CCMO-LLM and CCMO on DASCMOP1-3.

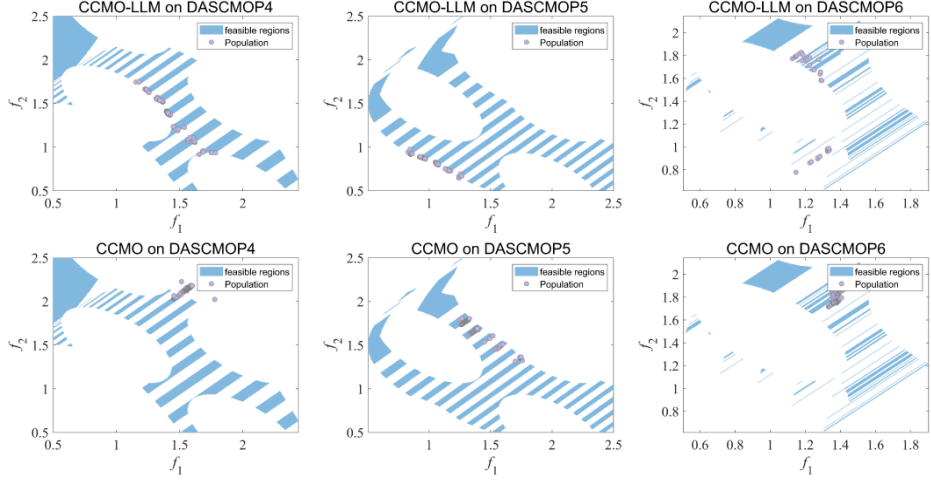

**Fig. 5.** The final population of CCMO-LLM and CCMO on DASCMOP4-6.

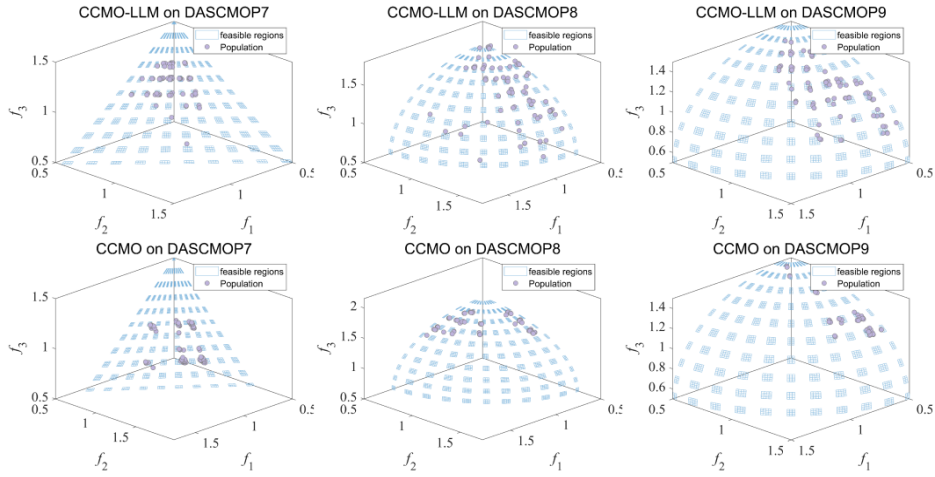

**Fig. 6.** The final population of CCMO-LLM and CCMO on DASCMOP7-9.

### 4.3    Results on the Real-world CMOPs

In this section, we evaluate the performance of the CCMO-LLM in addressing real-world CMOPs. We employ the RWMOP test suite, which is designed based on practical CMOPs and encompasses five distinct categories of problems from various fields. For a comprehensive assessment, we select a representative problem from each category for analysis: RWMOP5 (Disc Brake Design, Mechanical Design Problems), RWMOP22 (Haverly's Pooling Problem, Chemical Engineering Problems), RWMOP25 (Process Synthesis Problem, Process Design and Synthesis Problems), RWMOP30 (Synchronous Optimal Pulse-width Modulation of 3-level Inverters, Power

LLM-Aided Evolutionary Search for Constrained Multiobjective Optimization   13Electronics Problems), and RWMOP50 (Power Distribution System Planning, Power System Optimization Problems). The termination criterion is set to $FE_{max}$ = 10000. Table 3 presents the mean HV results obtained by CCMO-LLM and its five competitors on these RWMOP problems. The results clearly indicate that CCMO-LLM significantly outperforms the other five algorithms in solving these problems. Particularly, when compared to the original CCMO algorithm, CCMO-LLM, with the integration of LLMs, demonstrates enhanced problem-solving capabilities for RWMOP30 and RWMOP50. This outcome further substantiates the significant contribution of LLMs in generating offspring, thereby accelerating the convergence speed of the population.

Table 3. The average HV results obtained by CCMO-LLM and its five competitors on five real-world CMOPs.

| Problem | c-DPEA | CMME | MCCMO | MTCMO | CCMO | CCMO-LLM |
|---|---|---|---|---|---|---|
| RWMOP5 | 5.8444e-1 (3.59e-3) - | 5.8374e-1 (3.67e-3) - | 5.8747e-1 (3.27e-4) - | 5.8697e-1 (1.58e-3) - | 5.8462e-1 (2.79e-3) - | **6.1514e-1 (1.09e-2)** |
| RWMOP22 | NaN (NaN) | NaN (NaN) | **1.0751e+0 (1.69e-1)** | NaN (NaN) | NaN (NaN) | NaN (NaN) |
| RWMOP25 | 3.9471e-1 (1.63e-5) - | 3.9415e-1 (2.49e-4) = | 3.9467e-1 (3.58e-5) - | 3.9469e-1 (1.94e-5) - | 3.9469e-1 (3.63e-5) - | **3.9471e-1 (1.21e-5)** |
| RWMOP30 | NaN (NaN) | NaN (NaN) | NaN (NaN) | **3.8117e-1 (0.00e+0) =** | NaN (NaN) | 3.7351e-1 (3.01e-1) |
| RWMOP50 | NaN (NaN) | NaN (NaN) | NaN (NaN) | 4.3421e-2 (2.74e-3) = | NaN (NaN) | **4.4856e-2 (0.00e+0)** |
| +/-/= | 0/2/0 | 0/1/1 | 0/2/0 | 0/2/2 | 0/2/0 | ---- |

## 5 Conclusion

This study presents a novel approach that harnesses the power of LLMs in conjunction with traditional search operators to address the challenges posed by CMOPs. Through rigorous experimental validation, our proposed method has showcased significant advantages, outperforming the original CCMO algorithm across various scenarios and demonstrating robust competitiveness against other state-of-the-art CMOEAs. This research marks a pivotal step forward in integrating LLMs into the realm of CMOEAs, offering a promising avenue for advancing optimization techniques. By leveraging the strengths of LLMs to generate solutions that excel in both objective-based performance and constraint adherence, our approach represents a valuable contribution to the optimization community.

However, it is important to acknowledge that the application of LLMs still entails a notable time cost, despite the accelerated population convergence achieved. Looking ahead, future research endeavors could delve into exploring more efficient mechanisms for integrating LLMs with EAs. Additionally, the development of diverse combination strategies holds promise for further enhancing algorithm performance and application efficiency in real-world scenarios. In essence, this study not only sheds light on the potential of LLMs in optimizing complex problems but also lays the groundwork for future advancements in the field of evolutionary computation.